\documentclass{article}

  \usepackage[final]{nips_2018}
  
  \usepackage[utf8]{inputenc} 
  \usepackage[T1]{fontenc}    
  \usepackage{hyperref}       
  \usepackage{url}            
  \usepackage{booktabs}       
  \usepackage{amsfonts}       
  \usepackage{nicefrac}       
  \usepackage{microtype}      
  \usepackage[acronym]{glossaries}
  \usepackage[pdftex]{graphicx}
  \usepackage{caption}
  
  \newacronym{AD}{AD}{Alzheimer's disease}
  \newacronym{MRI}{MRI}{Magnetic Resonance Imaging}
  \newacronym{PET}{PET}{Positron Emission Tomography}
  \newacronym{CSF}{CSF}{Cerebrospinal Fluid}
  \newacronym{MMSE}{MMSE}{Mini Mental State Examination}
  \newacronym{SBMVSM}{SBMVSM}{Sparse Bayesian Multi-View Subject Model}
  \newacronym{ADNI}{ADNI}{Alzheimer's Disease Neuroimaging Initiative}
  \newacronym{SNP}{SNP}{Single Nucleotide Polymorphism}
  \newacronym{CN}{CN}{Cognitively Normal}
  \newacronym{CS}{CS}{Clinical Status}
  \newacronym{MCI}{MCI}{Mild Cognitive Impairment}
  \newacronym{IRT}{IRT}{Item Response Theory}
  \newacronym{ARD}{ARD}{Automatic Relevance Determination}
  \newacronym{AuC}{AuC}{Area Under The Curve}
  \newacronym{RMSE}{RMSE}{Root Mean Square Error}
  
  \title{Knowledge-driven generative subspaces for modeling multi-view dependencies in medical data}
  
  \author{
    Parvathy Sudhir Pillai\thanks{Preprint. Work in progress.} \\
    School of Computing\\
    National University of Singapore\\
    Singapore, 117417 \\
    \texttt{parvathy@comp.nus.edu.sg} \\
    \And
    Tze-Yun Leong\\
    School of Computing\\
    National University of Singapore\\
    Singapore, 117417 \\
    \texttt{leongty@comp.nus.edu.sg} \\
  }
  
\begin{document}
  
  \maketitle
  
  \begin{abstract}
    Early detection of \acrfull{AD} and identification of potential risk/beneficial factors are
important for planning and administering timely interventions or preventive measures.
    In this paper, we learn a disease model for AD that combines genotypic and phenotypic
    profiles, and cognitive health metrics of patients. We propose a probabilistic generative subspace 
    that describes the correlative, complementary and domain-specific semantics of the dependencies in multi-view, multi-modality 
    medical data. Guided by domain knowledge and
    using the latent consensus between abstractions of multi-view data, we model the fusion as
    a data generating process. We show that our approach can potentially lead to i) explainable clinical predictions and 
    ii) improved \acrshort{AD} diagnoses. 
   \end{abstract}
  \section{Introduction}
    \acrshort{AD} is a neurodegenerative disorder that may be influenced by many factors from genetic, medical 
    and family history, demographics and other personal attributes. 
    An \acrshort{AD} diagnosis is characterized by abnormalities in multiple diagnostic modalities including 
    neuroimages such as \acrfull{MRI} and \acrfull{PET}, and neuropsychological or clinical tests~\citet{Gray2012}.
    Thus, clinical data include markers from diagnostic modalities as well as protective/risk factors 
    from the patient's background. Each feature set provides a partial, yet different perspective
    to reveal the underlying cognitive health state of the study subject. With the recent progress in 
    multi-view machine learning, combinations of markers distinguish \acrshort{AD} patients from healthy
    controls with high accuracy~\citet{Weiner2017}. However, accurate detection at the early stages, 
    where interventions are most likely to be effective, remains challenging.
    Through data fusion, we seek to extract salient semantic representations of markers and background
    information, explain their dependencies, and improve key prediction metrics for \acrshort{AD} diagnosis
    and prognosis.
    
\begin{figure}
      \centering
      \begin{minipage}{.5\linewidth}
      \includegraphics[scale=0.3]{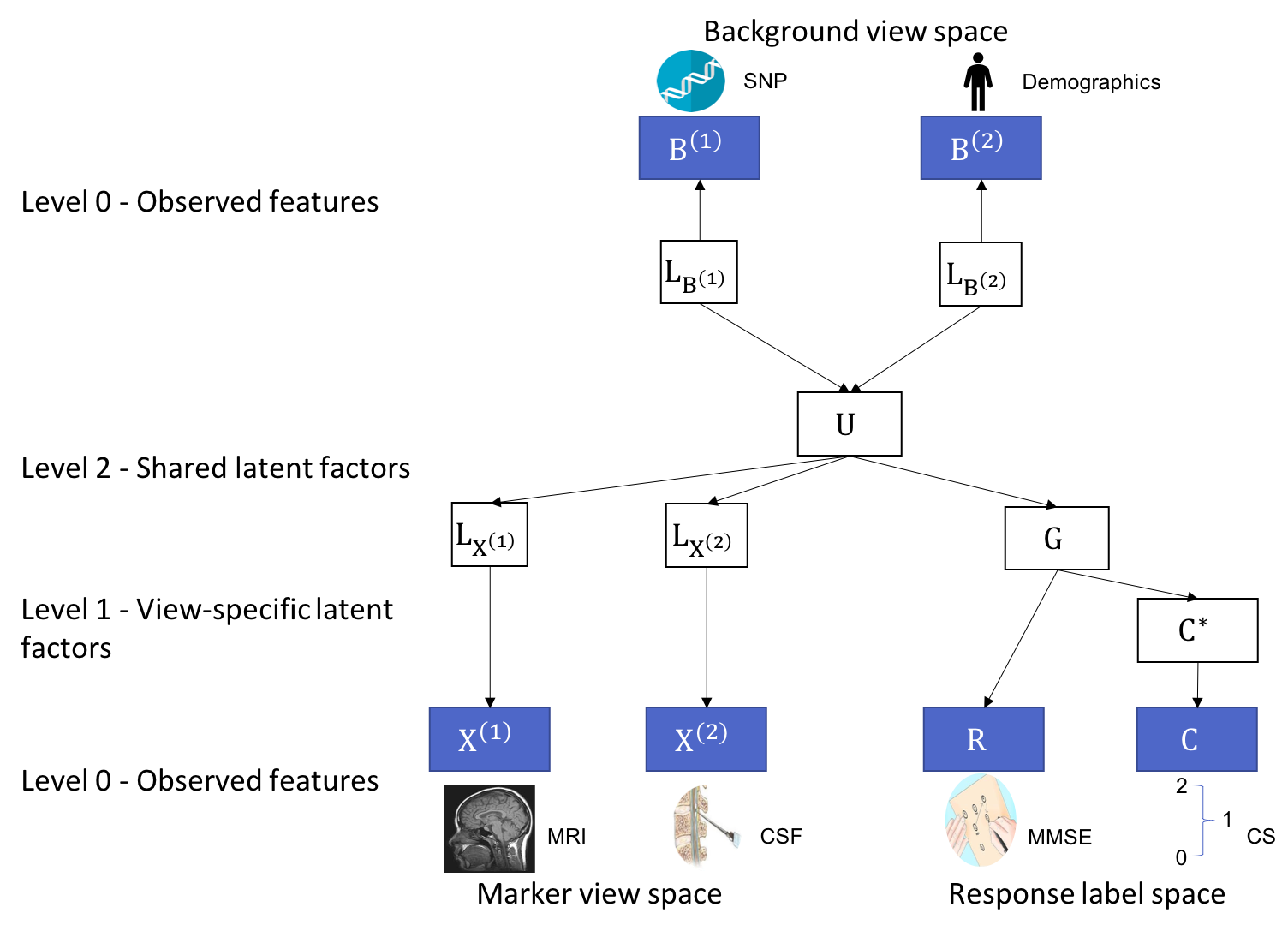}
      \caption{Sparse Bayesian Multi-view Subject Model}
      \label{fig:bmvl}
      \end{minipage}
      \hfill
      \begin{minipage}{.45\linewidth}
        \small
        \resizebox{\textwidth}{!}{%
        \begin{tabular}{p{2cm} p{1cm} p{1.5cm} p{1cm} p{1.5cm} p{1cm} p{1.5cm}}
          \toprule 
                & \multicolumn{2}{c}{\acrshort{AD}(n = 128)} &\multicolumn{2}{c}{\acrshort{MCI}(n = 287)}& \multicolumn{2}{c}{\acrshort{CN}(n = 174)}\\
                & Mean (SD) & Range 
                & Mean (SD) & Range  
                & Mean (SD) & Range  \\    	
        \midrule
        Age &75.4 ($\pm$7.3)	& 58.4-87.7	& 75.3 ($\pm$7.2)	& 55.1-88.8	& 75.2 ($\pm$5.2)	& 62-84.7\\
        \acrshort{MMSE} & 23.77 ($\pm$1.9)& 20-26	& 27.12 ($\pm$1.6) &24-30	& 29.04 ($\pm$1.2)&25-30\\
        \midrule
        & \multicolumn{2}{c}{66 M, 62 F} &\multicolumn{2}{c}{179 M, 108 F	CN} & \multicolumn{2}{c}{98 M, 76 F}\\
        \bottomrule
      \end{tabular}}
      \captionof{table}{Subject demography}
      \label{tab:demo}
    \end{minipage}
    \end{figure}

    We use a dataset of 589 subjects (refer to Table~\ref{tab:demo} for demography stats) from the 
    \acrfull{ADNI} dataset, which include their background features:
    demographic, genotypic (\acrfull{SNP}) and medical history, and markers: grey matter volumes from
    baseline \acrshort{MRI} and \acrshort{CSF} biomarkers, and cognitive measures: \acrfull{MMSE} and 
    \acrfull{CS} (\acrfull{CN}, \acrfull{MCI} and \acrshort{AD}). Each view has specific distributional properties; 
    \acrshort{MRI} volumes are continuous, medical history questions are categorical, clinical status is ordinal etc. 
    Also, features within a view could structurally be composed of several smaller features. 
    Probabilistic dependence between features could be mutual and non-directional (e.g., correlation) 
    or conditional and directional (e.g., conditional dependence/independence, causation).
    Differentiating the types of probabilistic dependence makes the model more intuitive and interpretable. 
    For instance, a person's cognitive capability assessed by \acrshort{MMSE} on a 30-point scale, consists of
    orientation, registration, recall, attention, language etc., which can be further decomposed. 
    The marks scored by a person in two questions may be correlated, but conditionally independent given 
    the clinical status.
    Therefore, background features influence the clinical condition which in turn decides the values of the markers.
    We learn the dependencies in multi-view clinical data by associating and generating them through their 
    conceptual abstractions. Subspace modeling relates measurable features (\emph{e.g.} \acrshort{MRI} volumes)
    to abstract concepts in a probabilistic way which allows flexible and efficient statistical reasoning and inference. 
    Shared subspaces are based on the idea of a latent consensus between views representing the same real-world entity.
    ~\cite{Zhe2015, Zhe2016, Lian2015} developed generative models to 
    model views of heterogeneous biomedical data through a shared latent subspace that represents the information 
    shared between the views.  
    ~\cite{Xue2017} recently proposed an optimization-based 
    approach that simultaneously learns individual information in each view but also captures feature correlations 
    among multiple views by learning the shared component. Though most of them enable supervised predictions, 
    we require an approach that simultaneously allows: i) view-specific dependencies, ii) shared dependencies
    between subsets of the views iii) inclusion of domain knowledge and, iv) multiple response variables.
    
    We propose a generative modeling of the probabilistic subspace that represents an individual's clinical condition
 as a latent continuum from healthy to being sick. We use view-specific subspaces to abstract
    concepts from markers (\emph{e.g.,} brain's structural integrity from volumetric \acrshort{MRI}),
    background (\emph{e.g.,} genetic profile from \acrshort{SNP}s) and cognitive measures 
    (\emph{e.g.,} cognitive capability from \acrshort{MMSE}) which are not measurable directly.
    We assume from domain knowledge that the dependencies between abstract markers and cognitive capability 
    are conditionally independent given the latent clinical condition, which is in turn conditioned on the 
    background feature abstractions in a hierarchical manner. A graphical representation of our approach,
    \textbf{\acrfull{SBMVSM}}, is depicted in Figure.~\ref{fig:bmvl}. The observed features at Level-0 comprise the views 
    constructed from markers, background and cognitive measures (colored cells in Figure.~\ref{fig:bmvl}). 
    Level-1 latent factors explain the view-specific correlations. Level-2 latent factors 
    represent the subspace shared across the view-specific latent factors. Through the model, we: 
    i) dimensionality reduction using sparse projections of the subspaces to the lower level, and
    ii) build the subject's clinical profile using continuous latent variables which are more efficient 
    at representing information, using fewer variables.
  \section{Model}
  \subsection{Notation:}
  We consider a supervised setting, where $\mathrm{X}^{(1)}\ldots\mathrm{X}^{(M_{mark})}$ represent the 
  the multi-view markers, $\mathrm{B}^{(1)}\ldots\mathrm{B}^{(M_{bg})}$, the background views, 
  $\mathrm{R}$, $\mathrm{C}$ the response variables \emph{i.e.,} \acrshort{MMSE} and clinical status \acrshort{CS} 
  respectively, $M_{mark}$ the number of marker views, $M_{bg}$ the number of background views 
  and $N$ the number of subjects.
  \subsection{Knowledge-driven generative modeling of the dependency subspace}
  \cite{Shwe1991, Pradhan1994, Seixas2014} translated the biomedical knowledge bases and datasets
  into multi-layered Bayesian networks. Each layer corresponds to specific types of variables: 
  i) layer 1 - predisposing (background) ii) layer 2 - diseases and iii) layer 3 - symptoms. 
  The layering represents the probabilistic relationships between different types of variables--
  the variables in layer 1 influence layer 2, which in turn affect layer 3.
  We follow a similar hierarchy to simulate the data generation process and include the domain knowledge
  from the epidemiology of \acrshort{AD}. We adopt the prior knowledge of a causal approach in which genetic variables like
  \acrshort{SNP}s and demographic variables such as age and sex are fixed before other variables and are
  not influenced by them. These should be \textit{roots} in the hierarchy, which impact the distributions 
  of the clinical condition. On the contrary, markers are dependent on the clinical condition and possibly
  background variables, but do not affect other variables and hence should be \textit{leaves} 
  in the hierarchy (~\citet{Jin2016}). We follow a full Bayesian treatment of \acrshort{SBMVSM}, 
  which includes the following steps:
  \paragraph{\textbf{Extracting view-specific subspaces: }}
    We use Bayesian matrix factorization~\citep{Nakajima2011} to extract latent subspaces from 
    continuous-valued views. For categorical-valued views, we use \acrfull{IRT}
    models~\citep{Hambleton1991} to express the observed variables as resulting from continuous latent traits,
    while for ordinal-valued views, we opt for the graded response model~\citep{Samejima1969} 
    to quantify the latent trait. The observed continuous views of markers, $\mathrm{X}^{(j)}$, and 
    background variables, $\mathrm{B}^{(k)}$, are linear transformations of uncorrelated
    low-rank latent factors, $\mathrm{L}_{\mathrm{X}^{(j)}}$ and $\mathrm{L}_{\mathrm{B}^{(k)}}$
    with $l^{(j)}$ and $l^{(k)}$ number of latent factors respectively
    (\emph{i.e.,}$\mathrm{L}_{\mathrm{X}^{(j)}_i} \sim \mathcal{N}_{l^{(j)}}(0, I_{l^{(j)}}$). 
    In a simplified sense, $\mathrm{X}^{(j)} = \sum_{r=1}^{l^{(j)}} \mathrm{L}_{\mathrm{X}^{(j)}} 
    \mathrm{V}^{(j)} + \epsilon_{\mathrm{X}^{(j)}}$, where $\mathrm{V}^{(j)}$ is the weight matrix and
    $\epsilon_{\mathrm{X}^{(j)}}$ is the noise. \acrshort{IRT} expresses the categorical views, $\mathrm{X}^{(j)}$ 
    (0 if `no'/1 if `yes') as resulting from a latent trait, $\theta_i^{(j)}$ ($\theta_i \sim \mathcal{N}(0,1)$) of sample, $i$,
    parameterized by the difficulty, $\delta_q$ (the slope of $P(X_{iq}^{(j)}  = 1|\theta_i)$) and discrimination, $\alpha_q$
    (where $P(X_{iq}^{(j)}  = 1|\theta_i) = 0.5$) of the corresponding feature, $q$ from $\mathrm{X}^{(j)}$ 
    We have the probability of a `yes' following a logistic function; $P(\mathrm{X}_{iq}^{(j)}=1|\theta_i,\alpha_q,\delta_q)$ = 
    $\frac{\exp{(\alpha_q(\theta_i-\delta_j))}}{1+\exp{(\alpha_q(\theta_i-\delta_q))}}$. 
    Similar equations apply for background views.
  \paragraph{\textbf{Generating shared subspace: }}
    We link the vertical concatenation of Level-1 marker abstractions, $\mathrm{L}_{\mathrm{X}}$ 
    through the Level-2 shared subspace, $\mathrm{U}$, using similar sparse projections. 
    We model $\mathrm{U}$ to follow a multivariate normal distribution with the mean expressed 
    as multiple linear regressions on the vertical concatenation of Level-1 background abstractions 
    $\mathrm{L}_{\mathrm{B}}$. 
  \paragraph{\textbf{Linking responses to shared latent subspace:}}
    The correlation between the continuous (e.g. \acrshort{MMSE}) and ordinal (e.g. \acrshort{CS}) response variables,
    $R$ and $C$ is captured using single continuous latent variable, $\mathrm{G}$. We first follow
    the principle that each ordinal variable, $C$, is a chopped-up version of a hypothetical
    underlying continuous variable ($C^*$) with a mean of 0. 
    Thus, $P(C_i|C^*_i) = {\sum_{c=0}}^2 \eta(C_i = c) \eta(b_c \leq C^* _i < b_{c+1})$.
    A patient, $i$ has a \acrshort{CS}, $c$, if $C^* _i$ falls between cutpoints, $b_c$ and $b_{c+1}$.
    Further, we model the two continuous response variables, $R$ and $C^*$ as projections of the
    latent variable, $\mathrm{G}$. 
    \paragraph{\textbf{Priors for weight matrices of projections: }}
    The view-specific weight matrices from Level-1, $\mathrm{V}^{(j)}$'s, have a horseshoe prior~\citep{Carvalho2009}, 
    to promote sparsity and reduce number of features loading on a particular factor. 
    Further, we use the group sparse \acrshort{ARD} priors~\citep{Tipping2001} for the shared subspace, $U$,
    to allow subsets of views if not all to interact. We also use appropriate hyperpriors to enable
    hierarchical Bayesian modeling.
  \section{Preliminary results}
  We build our inference engine in Stan probabilistic programming language~\citep{Carpenter2017, Stan2017}
  using its R interface~\citep{RStan2016}. All views share the the number of samples, $N$, as
  the common dimension. We represent each view as a matrix, $D^{(j)} \in \mathbb{R}^{N \times M^{(j)}}$, 
  where the rows correspond to the subjects and the columns to the features. 
  We utilize the fused representation from our \acrshort{SBMVSM} model to predict the clinical status 
  (2- AD, 1- MCI and 0-CN(cognitively normal)) and \acrshort{MMSE} scores. 
  Further, we decide on the dimensions of Level-1 ($\mathrm{L}_{\mathrm{X}}$, $\mathrm{L}_{\mathrm{B}}$) 
  and Level-2 ($\mathrm{U}$) subspaces through multiple cross-validations.
  
  In this study, we use the data collected from the \acrshort{ADNI} database~\citet{ADNI}. 
  The views we consider include \acrshort{MRI} volumes (90 features), 
  \acrshort{CSF} biomarkers (3 features), selected \acrshort{SNP}s (924 features)~\citet{Zhe2014},
  demographics (7 features). The resulting dataset after incomplete 
  data imputation/removal, consists of 589 study subjects (128 \acrshort{AD}, 174 \acrshort{CN}, 287 \acrshort{MCI}).  
  We use SMOTE oversampling~\citet{Chawla2002} to overcome class-imbalance and 
  and $z$-normalize all continuous features. We report the prediction performances over ten-fold cross-validation
  using the standard metrics of accuracy (Acc.), precision (Prec.) and recall (Rec.)
  for multi-class classification and, \acrfull{RMSE} for \acrshort{MMSE} score prediction.
  \begin{figure}
    \centering
    \includegraphics[scale=0.4]{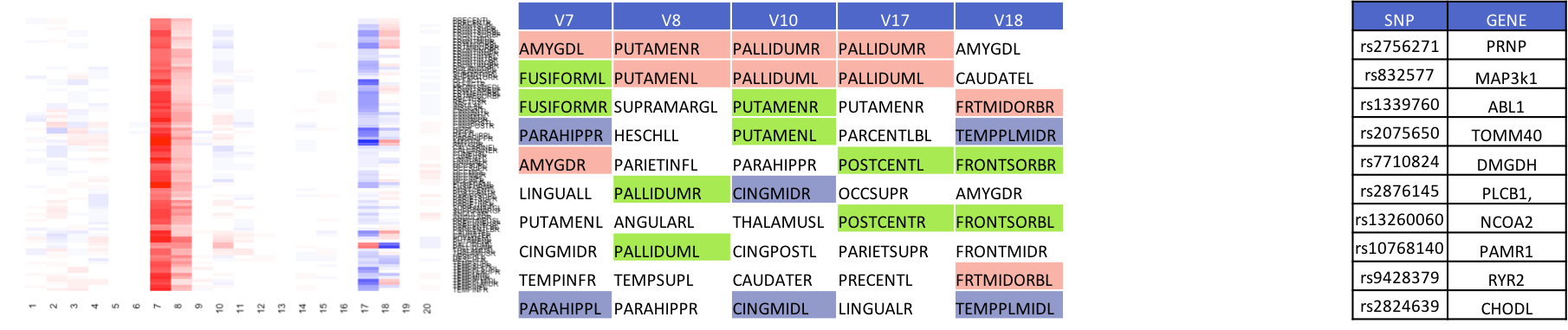}\\
    a)\acrshort{MRI} Level-1 factor loadings b)Top 10 \acrshort{MRI} features per factor \hspace{0.25cm} 
    c) Top \acrshort{SNP}s and Genes 
    \small\caption{View-specific factors of markers and background}
    \label{fig:results}
  \end{figure}
  We determine the dimensions of the best low-rank approximation of the \acrshort{MRI} view to be 20
  by cross-validation while predicting the clinical status. We present the weights
  of the Level-1 \acrshort{MRI} factors in a heatmap (Figure ~\ref{fig:results}a.). Further, we extract the 10 top-weighted features from
  the factors. A snapshot of the salient features from the factors with a higher say in clinical status 
  determination indicates that similar structures in the left and right brain fall under the same factor, due to their
  correlations(Figure ~\ref{fig:results}b.) e.g. AMYGDYL and AMYGDR in Factor V7.). View-specific latent 
  trait from \acrshort{SNP}s identify the ones in (Figure ~\ref{fig:results}c.) as the top 10. Many 
  of these belong to genes salient to \acrshort{AD} such as ABL1, MAP3K1, TOMM40 etc. 
  \begin{table}[h]
    \centering
    \small
    \resizebox{\textwidth}{!}{%
      \begin{tabular}{c|c|c|c|c}
        \toprule
        \textbf{Study} & \textbf{Modalities} & \textbf{N} & \textbf{Task} & \textbf{Prediction metrics}\\
        \midrule
        ~\citet{Zhu2016}& \acrshort{MRI}, \acrshort{PET} & 202 & & Acc. 76.4\\
        \acrshort{SBMVSM}& \acrshort{MRI}, \acrshort{PET} & 187 & \acrshort{AD} vs. \acrshort{MCI} vs. \acrshort{CN} & Acc. 79.09, Prec. 0.81, Rec 0.77\\
        \acrshort{SBMVSM}& \acrshort{MRI}, \acrshort{SNP}, \acrshort{CSF}, demographics & 589 & & Acc. 77.39, Prec. 0.78, Rec 0.74\\
        \midrule
        ~\citet{Zhu2016-2}& \acrshort{MRI}, \acrshort{PET} & 202 & & \acrshort{RMSE} 2.110 $\pm$ 0.41\\
        \acrshort{SBMVSM}& \acrshort{MRI}, \acrshort{PET} & 187 & \acrshort{MMSE} score regression & \acrshort{RMSE} 1.44 $\pm$ 0.38\\
        \acrshort{SBMVSM}& \acrshort{MRI}, \acrshort{SNP}, \acrshort{CSF}, demographics & 589 & & \acrshort{RMSE} 1.833 $\pm$ 0.35\\
        \bottomrule
      \end{tabular}}
      \caption{Prediction performance: Comparison with state-of-the-art}    
    \end{table}
  We fix the dimensions of the shared latent subspace, $\mathrm{U}$, as 20 through cross-validation. We perform multi-task predictions,
  clinical status classification and \acrshort{MMSE} regression by $\mathrm{U}$ to the latent response, $\mathrm{G}$.
  We present the results of the prediction tasks and compare with the state-of-the-art methods on the \acrshort{ADNI} dataset
  as presented in ~\citet{Weiner2017}. For a fair comparison, we use \acrshort{PET} and \acrshort{MRI} measures as 
  done by the works we compare against. Our model outperforms the state-of-the-art not only in prediction, but also
  interpretation of results through the depiction of multi-view dependence. \acrshort{SBMVSM} identifies correlations among features 
  by grouping them under the same factor and simulates the causal domain knowledge through the data generating process.
  Further, we use our method with \acrshort{MRI} and \acrshort{CSF} markers and, background 
  features such as \acrshort{SNP}, yet obtain comparable results as using specific neuroimage-based markers (\acrshort{PET} and 
  \acrshort{MRI}). 
  \section{Conclusion}
    We propose a probabilistic graphical model-based framework that takes into account the data generation 
    and dependency semantics among the features from disparate data sources. The generative framework identifies a 
    shared latent space between multiple marker and background views and response variables.
    \acrshort{SBMVSM} serves as a multi-view multi-task disease prediction model and achieves 
    good prediction performance in both the tasks in some preliminary studies.
  \small
  
  \bibliographystyle{apalike}
  \bibliography{Reference}
  
  \end{document}